\documentclass[
]{ceurart}

\usepackage{listings}
\usepackage{url}
\usepackage{hyperref}
\usepackage{setspace}
\usepackage{pgfplotstable}
\usepackage{float}
\usepackage{mathtools}
\usepackage{amsmath}
\usepackage{multirow}
\usepackage{adjustbox}
\usepackage{import}
\usepackage[shortlabels]{enumitem}
\usepackage{graphicx}
\usepackage{float}
\usepackage{lipsum} 
\usepackage{subcaption}
\usepackage{amsmath}
\usepackage{array}

\begin{document}

\copyrightyear{2022}
\copyrightclause{Copyright for this paper by its authors.
  Use permitted under Creative Commons License Attribution 4.0
  International (CC BY 4.0).}

\conference{Ontology Showcase and Demonstrations Track, 9th Joint Ontology Workshops
(JOWO 2023), co-located with FOIS 2023, 19-20 July, 2023, Sherbrooke,
Québec, Canada}

\title{Towards a Gateway for Knowledge Graph Schemas Collection, Analysis, and Embedding}



\author[1]{Mattia  Fumagalli}[%
email=mattia.fumagalli@unibz.it,
]
\address[1]{Free University of Bozen-Bolzano, Bolzano, Italy}
\address[2]{Qascom SRL, Vicenza, Italy}
\address[3]{University of Trento, Trento, Italy}

\author[2]{Marco Boffo}[%
email=marco.boffo@studenti.unitn.it,
]
\author[3]{Daqian Shi}[%
email=daqian.shi@unitn.it,
]
\author[3]{Mayukh Bagchi}[%
email=mayukh.bagchi@unitn.it,
]
\author[3]{Fausto Giunchiglia}[%
email=fausto.giunchiglia@unitn.it,
]



\begin{abstract}
One of the significant barriers to the training of statistical models on \textit{knowledge graphs} is the difficulty that scientists have in finding the best input data to address their prediction goal. In addition to this, a key challenge is to determine how to manipulate these \textit{relational data}, which are often in the form of particular triples (i.e., \textit{subject}, \textit{predicate}, \textit{object}), to enable the learning process. Currently, many high-quality catalogs of knowledge graphs, are available. However, their primary goal is the re-usability of these resources, and their interconnection, in the context of the \textit{Semantic Web}. 
This paper describes the \textit{LiveSchema} initiative, namely, a first version of a gateway that has the main scope of leveraging the gold mine of data collected by many existing \textit{catalogs} collecting relational data like ontologies and knowledge graphs. At the current state, LiveSchema contains $\sim1000$ datasets from 4 main sources and offers some key facilities, which allow to: \textit{i)} evolving LiveSchema, by aggregating other source catalogs and repositories as input sources; \textit{ii)} querying all the collected resources; \textit{iii)} transforming each given dataset into formal concept analysis matrices that enable analysis and visualization services; \textit{iv)} generating models and tensors from each given dataset.

\end{abstract}


\maketitle

\section{Introduction}

Finding the best data to train statistical models and properly address the target learning goal is widely recognized as one of the most pivotal tasks in \textit{Machine Learning} \textit{(ML)} \cite{geiger2020garbage}. ML models highly depend, indeed, on the quality of data they receive as input. While, so far, the development of highly efficient and scalable learning methods, to address critical prediction tasks (see, for instance, \textit{image classification} and \textit{information patterns recognition} \cite{anzai2012pattern}), helped data scientists and analytics professionals in scaling their activities, the process of finding, selecting and improving the quality of these data still requires a considerable amount of time and manual effort \cite{hastie2009elements}. This latter challenge is also present when statistical models are trained on \textit{knowledge representations}, like \textit{knowledge graphs} and \textit{ontologies} \cite{kejriwal2019knowledge}, where the data received as input are graph-structured data, consisting of \textit{entities} (or nodes) and labeled links, or \textit{edges}, (relations between entities). In this setting, the final learning goal can be identified as the \textit{prediction of missing relations between nodes}, the \textit{prediction of nodes properties}, and the \textit{clustering of the nodes based on their connections}, these being common goals arising in many scenarios, such as analysis of social networks and biological pathways \cite{nickel2015review}. Consequently, the efficacy of the ML algorithms directly depends on the quality of the input graphs, as well as their relevance to the domain of application. 

In this paper, leveraging the ideas presented in \cite{giunchiglia2016concepts,giunchiglia2017teleologies,giunchiglia2020entity} and \cite{giunchiglia2019knowledge}, where an approach to analyzing knowledge graph schemas to address \textit{Entity Type Recognition (ETR)} tasks has been devised, we introduce the \textit{LiveSche\-ma} initiative, namely the first version of a gateway that has the main scope of exploiting the gold mine of data collected by many existing \textit{catalogs} collecting relational data like ontologies and knowledge graphs. At the current state, LiveSchema contains $\sim1000$ datasets from 4 sources and offers some key facilities, which allow to: \textit{i)} continuously updating a catalog of knowledge graphs, by aggregating other source catalogs and repositories; \textit{ii)}  querying all the collected resources; \textit{iii)} transforming each given dataset into formal concept analysis matrices that enable analysis and visualization services; \textit{iv)} generating models and tensors from each given dataset. At the current state, LiveSchema is accessible at \url{http://liveschema.eu/} and it is ready to be demonstrated. The admin functionalities can be accessed and tested at \url{http://liveschema.eu/user/login}, by using \textit{`reviewer'} as \textit{admin/password}.

This demonstration paper is organized as follows. Section 2 introduces the motivation guiding the initiative. Section 3 gives a brief overview of the data architecture. Section 4 shows how LiveSchema can be evolved with new knowledge data. Section 4 illustrates the main LiveSchema components. Section 5 provides an example of how Liveschema can be used. Section 6 discusses some implications and limitations. Section 7 provides the conclusion.

\section{Motivation}

As an example scenario, suppose that a data scientist needs to run a standard \textit{Entity Type Recognition task (ETR)}\footnote{Relational data was proven to be key also in a \textit{transfer learning} setting \cite{fumagalli@ontology,fumagalli2019towards}.}, as it is described in \cite{giunchiglia2020entity} and \cite{sleeman2015entity}, where the goal is to recognize objects of the type `Person' across a set of multiple tabular data, coming, for instance, from an open data repository. This may involve that she needs to find a reference ontology containing: \textit{i.} the target class and corresponding label; \textit{ii.} possibly a huge number of properties for the target class, to increase the probability to match some of the input test properties; \textit{iii.} possibly a low number of overlapping properties, in order to decrease the number of false-negative/positive predictions.

The process of searching, analyzing, and transforming the target ontology can take a long time and it may involve a considerable effort. The scientist has, indeed, to go through a broad search over the available resources and related catalogs, possibly checking multiple data versions and formats. Moreover, once the candidate resources are identified, she should run an analysis of the data, to better understand their reliability concerning the target task. Additionally, this analysis (see, for instance, the simple data about the number of properties associated with each class) requires a processing phase that is assumed to be set up and run directly by the scientist. As a final step, if the scientist succeeds in finding the data she needs, a transformation process must be run to re-use the relational data in the reference ETR setup. \textit{What if the scientist can run all these operations in one single place with the support of ready-to-be-used built-in facilities?}

The idea of LiveSchema precisely arose from this key challenge. Firstly, the gateway aims at supporting scientists in better finding the relational data they need. Indeed, by leveraging the updates of some of the best state-of-the-art catalogs, LiveSchema should offer an aggregation service that allows searching and keeping track of the evolution of the knowledge representation development community in one place. 

Moreover, by implementing some key state-of-the-art libraries, LiveSchema aims at facilitating the data analysis and preparation process. Most of the implemented libraries, indeed, require an \textit{ad-hoc} set-up and may involve the combination of multiple components and environments, involving some coding and development skills that not all pure data scientists have. In this sense, LiveSchema aims at offering a platform that unites data analysis, data processing, and machine learning model deployment, making them easily accessible, reusable, and less time-consuming.

\section{Data Architecture}

The current version of LiveSchema is grounded in the \textit{CKAN}\footnote{https://docs.ckan.org/en/2.9/user-guide.html\#what-is-ckan} open-source data management system which is widely recognized as one of the most reliable tools for managing open data. We concentrate on the fundamental \emph{distinction} in CKAN which informs the data architecture of LiveSchema, namely that between \emph{dataset} and \emph{resource}\footnote{https://docs.ckan.org/en/538-package-install-docs/publishing-datasets.html}. A dataset is defined as a \emph{set of data} (e.g., BBC Sport Ontology) which may contain several resources representing the \emph{physical embodiment} of the dataset in different downloadable formats (e.g., BBC Sport Ontology in \texttt{TURTLE}, \texttt{FCA} formats). This distinction allows us, as a major advance from mainstream catalogs such as \cite{LOV}, to exploit \emph{fine-grained metadata} properties from the Application Profile for European Data Portals (DCAT-AP)\footnote{https://ec.europa.eu/isa2/solutions/dcat-application-profile-data-portals-europe\_en/}, which makes a \emph{conceptually identical} distinction between \emph{dataset} and \emph{distribution}. The additional advantage of using DCAT-AP is that it organizes metadata into \emph{mandatory}, \emph{recommended}, and \emph{optional} properties which are considered the \emph{key} for facilitating different levels of semantic interoperability amongst data catalogs. 

We now elucidate the metadata specification, i.e. the selected metadata properties for datasets and distributions considered for the current version of LiveSchema:
\begin{itemize}
\vspace{-0.5em}
    \item [i.] \textbf{Dataset}: 
    \begin{itemize}
        \item \texttt{MANDATORY}: \textit{description}, \textit{title};
        \item \texttt{RECOMMENDED}: \textit{dataset distribution}, \textit{keyword}, \textit{publisher}, \textit{category};
        \item \texttt{OPTIONAL}: \textit{other identifier}, \textit{version notes}, \textit{landing page}, \textit{access rights}, \textit{creator}, \textit{has version}, \textit{is version of}, \textit{identifier}, \textit{release date}, \textit{update}, \textit{language}, \textit{provenance}, \textit{documentation}, \textit{was generated by}, \textit{version}.
    \end{itemize}
    \item [ii.] \textbf{Distribution}: 
    \begin{itemize}
        \item \texttt{MANDATORY}: \textit{access url};
        \item \texttt{RECOMMENDED}: \textit{description}, \textit{format}, \textit{license};
        \item \texttt{OPTIONAL}: \textit{status}, \textit{access service}, \textit{byte size}, \textit{download url}, \textit{release date}, \textit{language}, \textit{update}, \textit{title}, \textit{documentation}.
    \end{itemize}
\end{itemize}

\noindent Notice that the distinction between dataset and distribution metadata is \emph{non-trivial} in the sense that metadata properties like \textit{format}, \textit{license}, \textit{byte size} and \textit{download url} are associated to a distribution and \emph{not} to the dataset itself.

Our first observation concerns the two \emph{major} advantages which the aforementioned data distinction and metadata specification brings to LiveSchema. Firstly, metadata enforces \emph{`FAIR'-ification} \cite{FAIR} of the KG schemas (which are `data' in this case), thus rendering them findable, accessible, interoperable, and reusable for the machine learning tasks which LiveSchema targets\footnote{To get an idea of what a FAIR catalog consists of, the work presented in \cite{barcelos2022fair} and accessible at \url{https://github.com/OntoUML/ontouml-models} provides a key reference}. Secondly, as a consequence of the first advantage, the metadata-enhanced KG schemas also play a pivotal role in initiating, enhancing, and sustaining \emph{reproducibility} \cite{RCR} which is \emph{key} for LiveSchema vis-à-vis the target machine learning ecosystem in which it participates.

Our second observation concerns the future extensibility of the metadata specification of LiveSchema. The starting distinction between dataset and distribution can help bootstrap the extension of the initial metadata specification to \emph{ontology-specific} metadata which, \emph{mutadis mutandis}, preserves the same distinction via the notions of \emph{ontology conceptualization} and \emph{ontology serialization} \cite{OMV}. One of the \emph{key} advantages of using ontology-specific metadata in LiveSchema is that the user can perform a highly customized (conjunctive) search, for instance, even at the level of logical formalism or ontology design language, thus retrieving the most compatible schema for the machine learning task at hand. In this direction, we plan to exploit the MOD \cite{MOD} ontology metadata proposals in the immediate future.

\section{Evolving LiveSchema}

\subsection{Data Collection and Development}
At the current state, LiveSchema relies on four main state-of-the-art catalogs, namely LOV, DERI\footnote{\url{http://vocab.deri.ie/}}, FINTO\footnote{\url{http://finto.fi/en/}} and a custom catalog which is still under construction\footnote{\url{http://liveschema.eu/organization/knowdive}}, where some selected resources are stored. 

Each catalog is associated with a provider, which is the person or organization that uploaded the data set and is in charge of its maintenance in the source catalog. From each catalog, multiple data sets have been individually scraped and uploaded in an automated way. Currently, LOV is the catalog providing most of the data sets, being one of the most widely used catalogs of vocabularies in the semantic web context. The extension of LiveSchema with new catalogs is part of the immediate future work. 

Given the selected catalogs, 26 types of metadata have been scraped, namely: \textit{.i} \textbf{Catalog}: \textit{id}, \textit{name}, \textit{title}, \textit{logo}, \textit{URL}, \textit{description}; \textit{.ii} \textbf{Data-Set}: \textit{id}, \textit{name}, \textit{title}, \textit{notes (description)}, \textit{issued}, \textit{modified (last modified)}, \textit{language}, \textit{uri (landingPage)}, \textit{contact-uri (homepage / contactPoint)}, \textit{maintainer / publisher (Provider)}, \textit{author / creator (Provider)}, \textit{license-id (license)}, \textit{license-url (license)}, \textit{owner-org (Catalog)}, \textit{version (versionInfo)}, \textit{tags (keyword)}, \textit{source}; \textit{.iii} \textbf{Provider}: \textit{id}, \textit{name}, \textit{title}, \textit{uri}.

We carefully checked the dataset during data scraping to ensure that LiveSchema is not breaking any license agreement. Currently, five kinds of licenses are admitted given their restrictions (all of them are part of the  \textit{Creative Commons}\footnote{\url{https://creativecommons.org/}} initiative). These license constraints need to be checked since we both provide access and we manipulate their content to provide the following resources. 
As we parse them from their source from various sets of formats\footnote{\url{https://rdflib.readthedocs.io/en/stable/plugin_parsers.html}}, we serialize them into the most common ones, namely \texttt{RDF} and \texttt{Turtle}. More advanced output formats can be generated through the processing operations enabled by the LiveSchema services, namely \texttt{CSV} (where all the triples and metadata of the input relational data are stored in a datasheet format), \texttt{CUE} (where all the cue metrics are provided), \texttt{FCA} (i.e., the FCA transformation matrix result), \texttt{VIS} (the format that can be used to enable visualization services functionalities), \texttt{EMB} (the format used to generate a statistical model based on a knowledge embedding process). 

\subsection{Stoking LiveSchema}
LiveSchema is managed by a group of knowledge experts, software engineers, and data scientists that contribute to the development and evolution of the whole system. This group of experts, whom we call here \textit{LiveSchema administrators}, or simply \textit{admins}, besides handling maintenance issues, are in charge of applying the evolution component functionalities. These functionalities are playing central roles in the process of populating the catalog. Two evolution operations can be applied by the LiveSchema admin. Firstly, LiveSchema provides an automated evolution process, which is composed of a parsing phase and a scraping phase. Few checkpoints are released for administrators to supervise the output of the automatic processes. Secondly, manual datasets uploading, reviewing, and managing are also available through the usage of LiveSchema services.

An example of the manually created list, containing new useful data sets, which are not present in the other selected catalogs, with all their relative information and metadata, is accessible\footnote{\url{https://github.com/knowdive/resources/blob/master/otherVocabs.xlsx}}. Some of these data sets are not directly obtainable from the web and they had to be downloaded, unzipped, or edited, and then uploaded on GitHub in order to render them collectible using an URL link. Once the data is scraped, a second key semi-automated parsing process is applied.

The parsing process is very simple, it is executed iteratively and has the goal of producing two main outputs, namely a set of serialized data sets and a set of parsed data sets. The first output is produced by scanning the data sets list and parsing it using RDFlib python library\footnote{\url{https://github.com/RDFLib/rdflib}}, namely a library that is used to process RDF resources. Here the produced output is used to generate more standard reference formats, which, in the current setting are represented by \texttt{RDF} and \texttt{Turtle}. We also allow for the generation of an \texttt{xlsx} (or \texttt{csv}) file encoding all the information (e.g., triple and metadata) about the data set to easily enable the other applications provided by the catalog. In this step, the key role of the admin in charge of the parsing process is to edit the data set list in order to filter out undesired data sets and parse only the ones that are required. The second output is produced by scanning each triple of the input data set. The filtering among the predicates to specify the focus of the dataset is applied just before the application of some services.

 \begin{figure}[!h]
   \centering
   \includegraphics[width=0.7\linewidth]{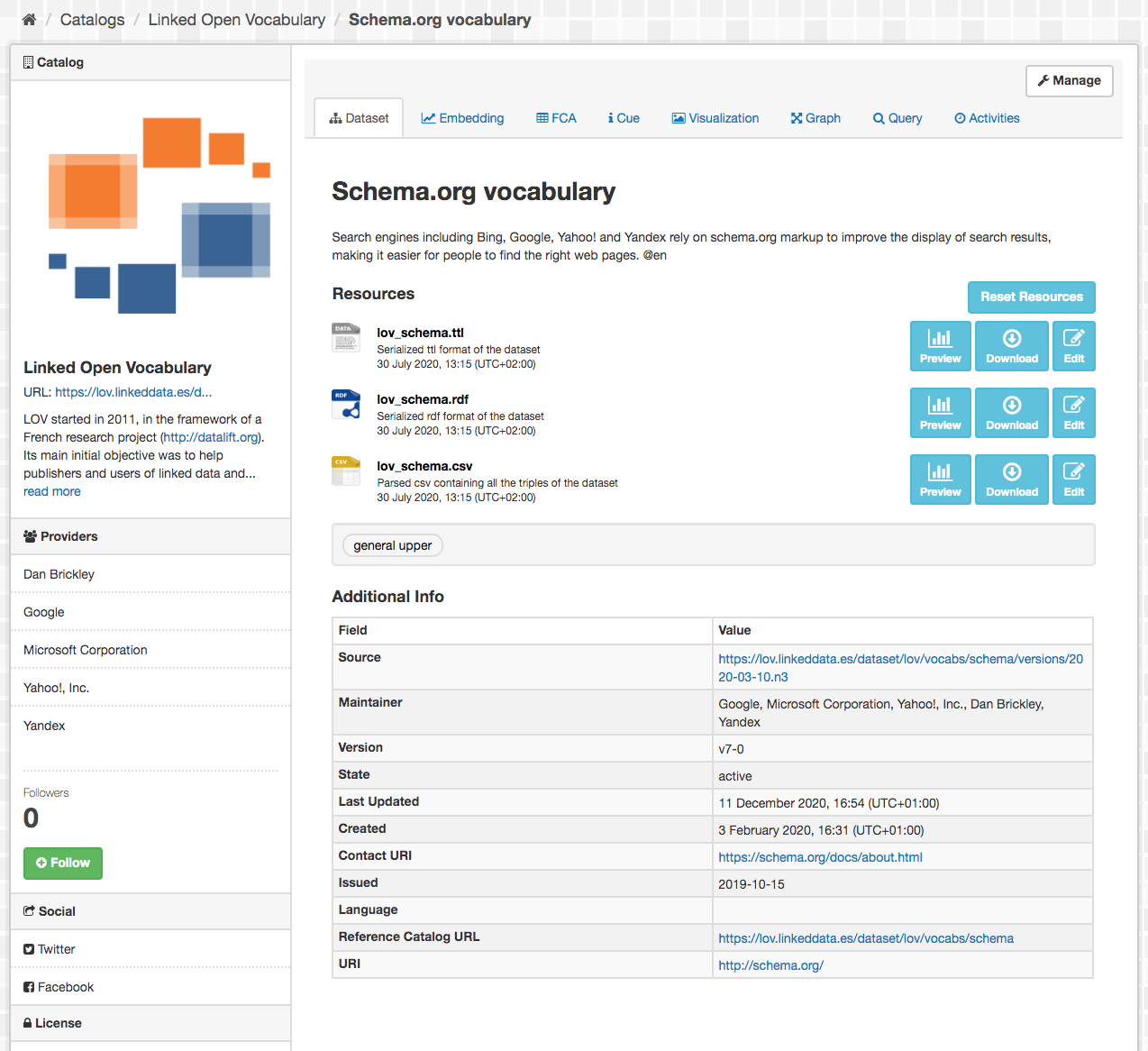}

   \caption{LiveSchema data set page (from an admin perspective).}
   \label{01}
\end{figure}

\subsection{Forging Datasets}
All the datasets that are gathered from the source catalogs and uploaded to LiveSchema can be then transformed and used as input of the available functionalities. The current LiveSchema version contains six main functionalities, which are 1) \textit{FCA generator}, namely the process by which data can be converted in the FCA format (\texttt{FCA}); 2) \textit{CUEs generator}, i.e., the process by which the CUEs (as defined in Section 2) are generated and encoded in the \texttt{CUE} format; 3)  \textit{Visualization generator}, namely the process by which the input data can visualized and analyzed (see \texttt{VIS} format; 4) \textit{Knowledge Embedder}, i.e., the application by which a model can be created out of the input data, by applying one or some of the libraries provided by the PyKEEN package\cite{ali2021pykeen} (see \texttt{EMB} reference format)\footnote{\url{https://pypi.org/project/pykeen/}}; 5) the \textit{Query Catalog} service, which allows running SPARQL queries\footnote{\url{https://rdflib.readthedocs.io/en/stable/intro_to_sparql.html}}; 
6) the \textit{knowledge graph visualizer}, namely an implementation of the WebVOWL\footnote{\url{http://vowl.visualdataweb.org/webvowl.html}} library. This set of functionalities can be easily accessed and reused utilizing APIs services, and can also be easily extended, e.g., 4), 5) and 6) can be run by directly using \texttt{.rdf} files as input. Each functionality may require an \textit{ad hoc format} to produce the output, and, in some cases, it may have some dependencies with the input format of other functionalities, e.g.,  1), 2) and 3) involving new formats. 


Figure \ref{01} above provides an example of the LiveSchema data set management, where all the available functionalities for managing, analyzing, and transforming the data are presented\footnote{three current reference resources \texttt{.rdf}, \texttt{.ttl} and \texttt{.csv} are ready to be downloaded.}. A set of metadata, tags, and information about the reference source catalog are also provided to users on the top left of the link. All the new formats (if present) are accessible on the corresponding functionality page. 

\section{LiveSchema Components}

\begin{figure}[!h]

  \centering
  \includegraphics[width=0.6\linewidth]{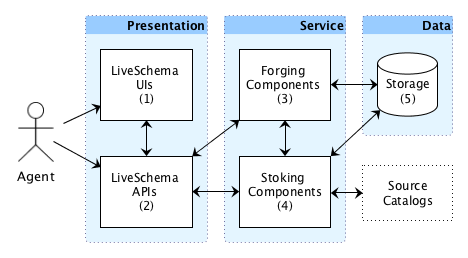}

  \caption{Overview of the LiveSchema components.}
  \label{02}
  \vspace{-0.5em}
\end{figure}

The main components of LiveSchema are combined as in Fig. \ref{02}. LiveSchema includes five main components (See Figure \ref{02}): (1) \textit{user interfaces UIs}; (2) the \textit{APIs}, provided by the CKAN platform, and that we partially customized according to our set-up needs, provide the main accesses to the LiveSchema environment; (3) \textit{stoking components}; (4) \textit{forging components} cover the main novel contributions of the LiveSchema initiative and offer the possibility to harvest, generate and process data; (5) \textit{Storage} allows to collect in one place all the data collected by other catalogs, provided by the users or generated through the services. All these components are grouped into three main layers: the \textit{presentation layer}, the \textit{service layer}, and the \textit{data layer}.  

\paragraph{Presentation layer.} This layer enables a community of users: \textit{.i} to maintain the whole gateway and its applications, and \textit{.ii} to suggest and upload new resources or edit some already existing resources. LiveSchema is mainly managed by a group of expert knowledge engineers, software engineers, and data scientists that contribute to the development and evolution of the whole catalog. Moreover, a group of guest users can also be involved in the collaborative development of the storage, by uploading and editing some new data sets and, possibly, creating new input reference catalogs, following well-founded guidelines provided by the knowledge engineers that administrate the ecosystem. The definition of the types of access and the different roles played by the LiveSchema users is part of the immediate future work. The APIs allow the users to exploit all of the website's core functionalities by external code. Using the API, developers will be able to: get JSON-formatted lists of vocabularies, with providers (namely, the agent who created the data set), source catalogs, or other LiveSchema information; get a full JSON representation of a data set or other related information derived from the analysis of the data set; search for data sets, providers, or other resources matching a query; create, update and delete data sets, with related metadata and information; get an activity stream of recently changed data set on LiveSchema, obtaining also the versioning information of each resource The UIs allow users to access data and functionalities. Two types of user interfaces are present, namely front-end and back-end user interfaces. The former is a customization of the standard CKAN template, where the home page allows to access all the contents of the website through five main widgets: 1) a menu with the top-level categories of the catalog, 2) a search form to easily access and browse data sets, 3) a showcase of the top services and a list of the source reference catalogs, 4) recent activities. Differently, the back-end user interface can be accessed only with credentials and allows for the editing and submission of existing or new data, or it enables the usage of some more applications. 

\paragraph{Service layer.} The stoking components are mainly necessary to check and gather any new knowledge resources from a set of previously selected catalogs. LiveSchema mainly relies on manual processes and a semi-automated process for data insertion. The former can be applied by any type of user, by submitting a new resource through a dedicated panel, but requires a review process from the administrator users. The latter is applied by selecting source catalogs as input and can be used to keep track of their updates. This stoking facility can be primarily customized by determining how many times the source catalogs must be checked and by defining what types of data sets can be collected and uploaded into the main storage. Currently, the quality criteria to allow the uploading of a data set, is the size, the type of license, and the correct format of its content. Along with the stoking components, the LiveSchema forging components encode a set of functionalities, which are aimed at the analysis and transformation of data, and the generation of new formats. All these functionalities are aimed at supporting scientists in the re-usage of the selected relational data. 

\section{Using LiveSchema}

The scope of this section is mainly to show how the LiveSchema processing component works. Through a running example, we illustrate how a user can exploit main functionalities. All the described operations can be directly tested by exploring and using the LiveSchema ecosystem, which is accessible at \url{http://liveschema.eu/}. The admin functionalities can be accessed and tested at \url{http://liveschema.eu/user/login}, by using \texttt{`reviewer'} as \textit{admin/password}.

\subsection{Analyzing Relational Data}

As an example scenario, suppose we need to run a standard \textit{entity type recognition task}, as it is described in \cite{giunchiglia2020entity} and \cite{sleeman2015entity}, where we may need to recognize objects of the type `Person' across a set of multiple tabular data, coming, for instance, from an open data repository. This may involve the need to find a reference relational model with \textit{.i} the target class and corresponding label; \textit{.ii} possibly a huge number of properties for the target class, in order to increase the probability to match some of the input test properties; \textit{.iii} possibly a low number of overlapping properties, in order to decrease the number of false negative/positive.

A LiveSchema user can perform a simple search across the available data sets that are present in the catalog and then run an analysis to select the best. The LiveSchema search facility exploits the CKAN search engine that allows for a quick ‘Google-style’ keyword search. All the data sets, providers, and group fields are searchable and the users can use all of them to research the desired entity. Thanks to this search functionality it is possible to provide a complete and customized service to the scientist looking for the desired ontology. The basic supported search options are \textit{.i} search over all the data sets attributes, namely by using any of the applied metadata; \textit{.ii} full-text search; \textit{.iii} fuzzy-matching, namely an option to search for closely matching terms instead of exact matches; \textit{.iv} search via API. Now, suppose that the user identifies three candidate resources for the goal ETR task, namely \textit{Schema.org}\footnote{https://schema.org/} (reference standard to support the indexing of web documents by \textit{Google}\footnote{https://www.google.com/}), \textit{FOAF}\footnote{http://xmlns.com/foaf/spec/} (a widely used vocabulary in the context of social networks) and the \textit{BBC sport ontology}\footnote{https://www.bbc.co.uk/ontologies/sport} (the ontology used by the BBC to model supports events, and roles). The next step is to access each single data set and check its meta-information, which can be done by first generating the \texttt{FCA} format for the selected resources.


Each LiveSchema data set has a dedicated page collecting its information, and where each processing functionality can be accessed. Here information about the related source catalog is provided as well, and the available standard reference formats can be downloaded. The FCA functionality can be accessed through the corresponding tab and allows for the generation of the corresponding matrix for each given input relational model. On the FCA service page is also possible to customize the generation of the matrix by filtering the target predicates. Then, multiple insights can be extracted by using the functionalities represented by each tab on the data set page. By downloading all the cue information comparisons between the three representations of `Person', provided by each ontology can be run. Table 1 represents the cue values for the given resources. From a quick benchmark is clear that in Schema.org, even if the cue of Person is not at the top, the given class has a high centrality with a score of 23. 

\begin{figure}[ht]
\centering
\begin{subfigure}[]{0.4\linewidth}
\small
\centering
\begin{tabular}{ccc}
\hline
\textit{Class} & \textit{$Cue_{e}$} & \textit{$Cue_{er}$} \\ \hline
`Person' - Schema.org & \textbf{23} & \textbf{0.81} \\
`Person' - FOAF & \textbf{3} & \textbf{0.82}\\
`Person' - BBC & \textbf{0.73} & \textbf{0.75} \\ \hline
\end{tabular}
\vspace{1em}
\end{subfigure}
\hspace{5mm}
\begin{subfigure}[]{0.4\linewidth}
\includegraphics[width=1\linewidth]{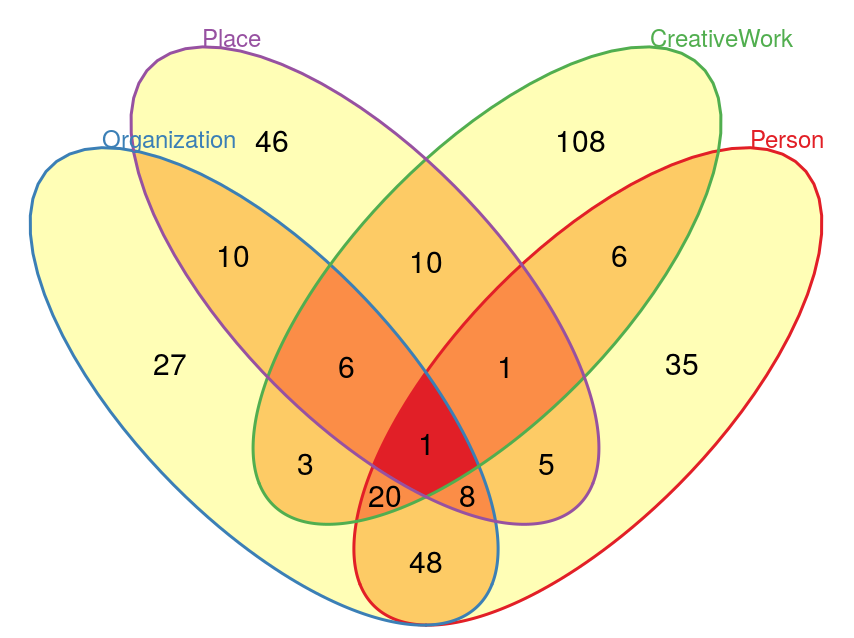}
\caption{}
\end{subfigure}
\caption{Cue values for the class `Person'.}
\label{double}
\vspace{-0.5em}
\end{figure}

Besides the quantification of the cues, further analysis can be run by visualizing the intersection of some of the top classes of the given resources. Figure \ref{08} represents an example of \textit{knowledge lotus} that can be extracted by the input resources. Knowledge lotuses are \textit{venn diagrams} that can be used to focus on specific parts of the input resources and they are particularly useful to represent the diversity of classes in terms of their (un-)shared properties. 
The yellow petals of the lotus show the number of properties that are distinctive for the given class. In the example, Person has 35 un-shared properties. The different shades of orange represent the number of properties shared with other classes (for instance, there is 1 property that is shared with all the classes).  

\begin{figure}[!t]
  \centering
  \includegraphics[width=0.7\linewidth]{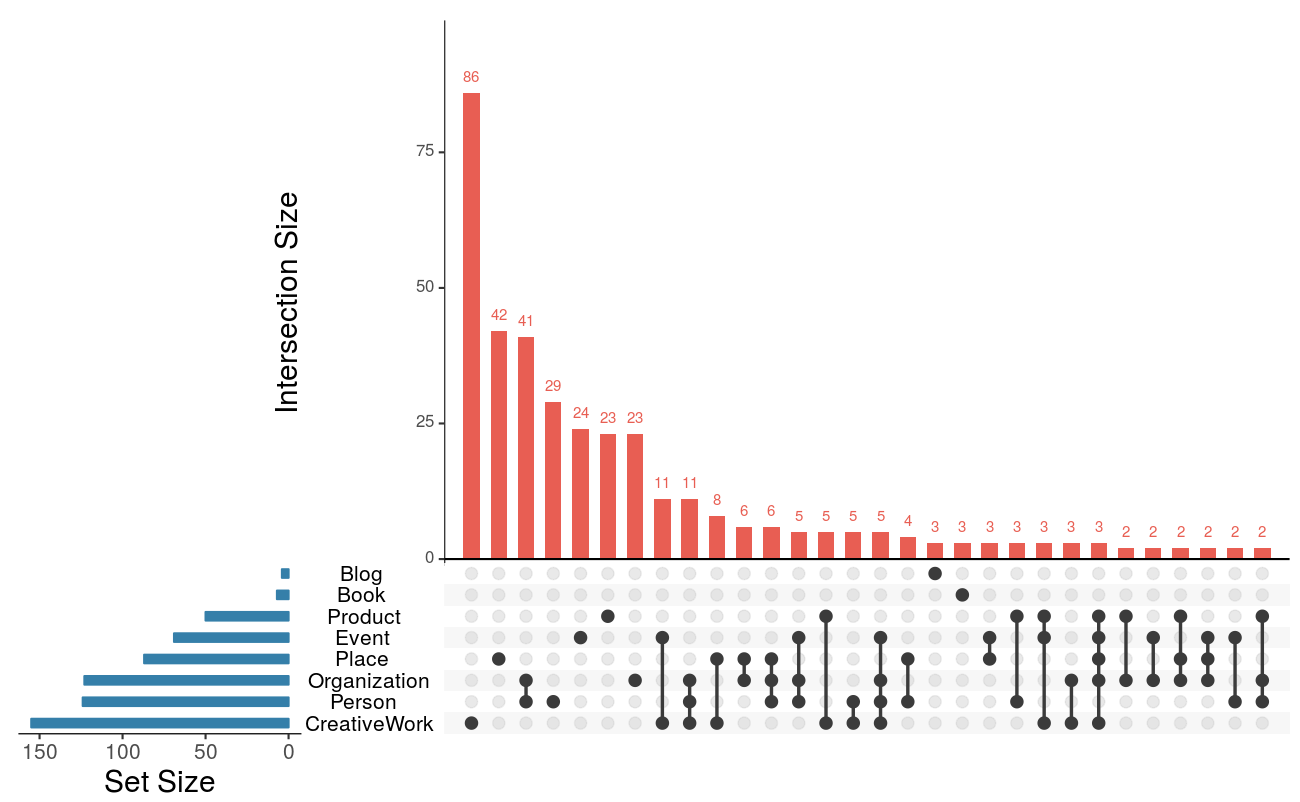}
  \caption{Etypes properties intersection: the UpSet visualization}
  \label{09}
\end{figure}

Further analysis can be run by applying the UpSet (multiple set) visualization facilities, which allows us to analyze the intersections between classes, by selecting more than 6 sets (the limit for knowledge lotuses). LiveSchema allows for both knowledge lotuses and UpSet visualization by embedding the functionalities of the intervention visualization environment\footnote{https://intervene.shinyapps.io/intervene/}. This environment was created for the visualization of multiple genomic regions and gene sets (or lists of items).
The main goal of the provided visualization options is to facilitate the analysis and interpretation of the input resource. An illustrative example of the representation of a resource utilizing the UpSet module is provided by Figure \ref{09}. Here 8 classes are selected. The blue bars on the left show the size of the classes in terms of the number of properties. The black dots identify the intersections between the classes and the red bars on top f the figure shows the size of the properties intersection set.  

\subsection{Embedding Relational Data}

Once the scientist has selected her resource, she is ready to embed it and generate a statistical model out of it. Notice that, in the current release of LiveSchema we allow for distributional embedding techniques only. The implementation of symbolic approaches, such as Inductive Logic Programming for addressing new tasks like, for instance, class expression learning, is part of the immediate future work. In this current setting, LiveSchema relies on a recent library collecting most of the state-of-the-art techniques for graph embedding, namely the PyKEEN library \cite{ali2021pykeen}. PyKEEN is a widely used solution for generating custom embedding models. It allows selection across a wide range of training approaches with multiple parameters and will output a \texttt{.pkl} file which can be directly imported inside ML pipelines.

\begin{figure}
  \centering
  \includegraphics[width=0.8\linewidth]{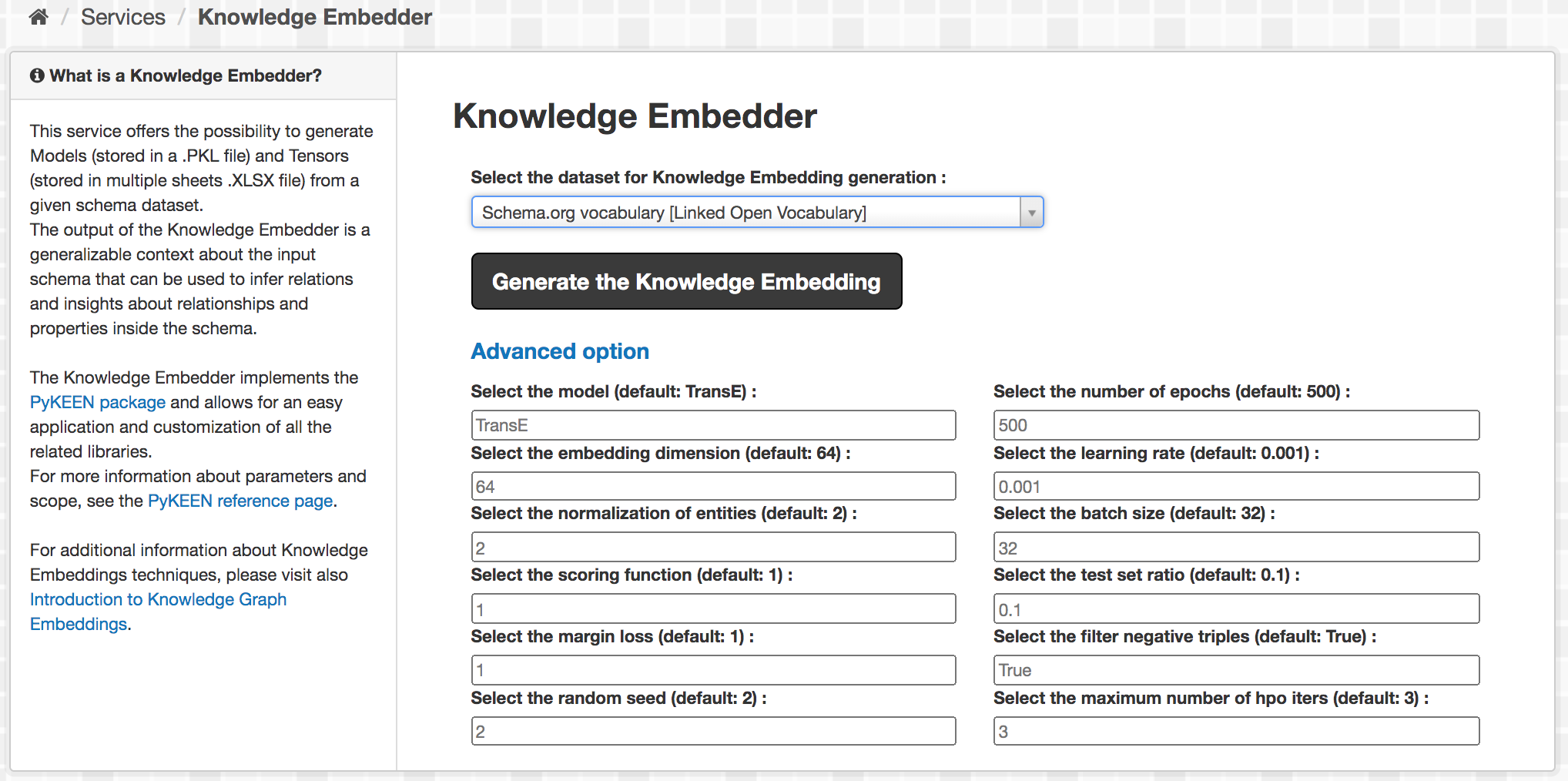}
  \caption{The Knowledge Embedding interface in Liveschema (zoom in for more details).}
  \label{10}
  \vspace{-0.5em}
\end{figure}

Figure \ref{10} demonstrates a screenshot of the LiveSchema KnowledgeEmbedder page, various parameters can be selected to obtain the specific learning goal. We can select the ``embedding model'' where we can select state-of-the-art algorithms, like \textit{TransE}, \textit{RESCAL} or \textit{DistMult} \cite{wang2017knowledge}; and settings like the ``loss function'', which is typically used to minimize the error of the model and can be used for reducing multiple features of the models to a single number, namely a scalar value, which allows candidate solutions to be ranked and compared \cite{schmidhuber2015deep}.

Notice that in LiveSchema we have data sets encoding relational models with no instance data (e.g., we have the DBpedia schema, but we do not have the so-called ABOX). This did not prevent us to adapt the embedding process and focus on the schema level only (relying on relational data we always have, indeed, triples: \textit{heads}, \textit{tails}, and \textit{relations}). This, besides opening the possibility to test a new application scenario, does not exclude the possibility to apply the standard approach where populated schemas are used as input. The population of LiveSchema with this kind of data is part of the immediate future work.

 

\section{Discussion}
We believe that LiveSchema could be a useful support to study relational data for both knowledge representation tasks (e.g., designing a knowledge base for enabling the interoperability between systems), and machine learning tasks, in particular the ones that rely on relational data structures for training their models. In this section, we discuss the implications of the initiative. Moreover, by identifying the limitations of our current setting, we also discuss opportunities for future work.

\subsection{Implications}
Firstly, LiveSchema can support scientists in finding the relational data they need. By leveraging the updates of some of the best state-of-the-art catalogs, LiveSchema can offer an aggregation service that allows keeping track of the evolution of the knowledge representation development community in one place. Notice that this does not aim to substitute the function of each single source catalog. The scientist can indeed access the source catalog and related \textit{ad hoc} services, if needed, directly from the LiveSchema gateway, this being also an opportunity of increasing the visibility of the vocabularies themselves.  

Another key point is that LiveSchema can represent an opportunity to bridge the gap between two key artificial intelligence communities, namely the knowledge representation and the machine learning community. While most of the data that are present in LiveSchema are indeed in a format that is compliant with the knowledge representation applications requirements, each data set can be also transformed so that it can be easily employed in machine learning set-ups. The analysis and embedding facilities offer further support in this direction. We believe that this is a way of supporting the exploitation of the huge amount of work done by the community and of making the relational model more accessible to machine learning scientists. 

Moreover, we implement state-of-the-art libraries to support data scientists in the data analysis and preparation phases. Most of the implemented libraries require coding and development skills, which will limit the usage of data scientists. To solve this issue, LiveSchema offers a platform that unites data analysis, data preprocessing, and machine learning model deployment, which makes them easily accessible and usable.

Finally, the overall project was also devised to pave the way for large case studies. Integrating knowledge representation and machine learning scenarios may indeed be devised in a different way of designing relational structures, with a different focus on some of their features or constraints (e.g., the number of properties to be used for describing a class or the overlapping between classes). Moreover, data scientists, reusing relational models for their predictive tasks, may better realize what relational models can be better than others about the specific learning target, and how they should be tuned to better support their task.

\subsection{Limitations}



Developing LiveSchema as a community of data scientists that exchange and reuse data to the benefit of the AI community is our long-term objective and this triggers the agenda for immediate future work. To achieve this goal, with the current set-up, there is still a gap that needs to be bridged.

As long as the LiveSchema observatory will grow, serious challenges about the scalability of the approach still need to be handled. One issue is that through the current version of the evolution component is not possible to automatically check duplicated resources coming from different vocabularies. 
Another pending issue, which is part of the agenda for the immediate future work, is the definition of a processing component functionality that enables users to work with multiple data sets together, this would be an important option, especially for supporting data integration tasks and the evolution of more robust machine learning models. A possible way of implementing this functionality will be to develop a new version of the current FCA conversion process, where multiple data sets can be given as input and then merged, by computing their similarities, into one single file. The output file will be used as a single resource containing the information of all its component datasets.

\section{Conclusion}

In this paper, we introduced the first version of the \textit{LiveSchema} gateway, which aims at exploiting the relational representation of ontologies not only for their classical application but also for their use in machine learning scenarios. The long-term goal of LiveSchema is to leverage the gold mine of data collected by many of the existing relational knowledge representations catalogs and offer a family of services to easily access, analyze, transform, and re-use data, with an emphasis on relational machine learning pipelines setup and predictive tasks.

\begin{acknowledgments}
The research conducted by Mattia Fumagalli is supported by the \emph{``Dense and Deep Geographic Virtual Knowledge Graphs for Visual Analysis - D2G2''} project, funded by the \emph{Autonomous Province of Bolzano}. The research conducted by Fausto Giunchiglia and Mayukh Bagchi has received funding from JIDEP - under grant agreement number 101058732. The research conducted by Daqian Shi has received funding from the program of China Scholarships Council (No. 202007820024).
\end{acknowledgments}

\bibliography{sample-ceur}

\appendix

\end{document}